\pdfoutput=1

\documentclass[11pt]{article}

\usepackage{acl}

\usepackage{times}
\usepackage{latexsym}

\usepackage{graphicx}
\usepackage{amsmath}
\usepackage{booktabs}
\newcommand{\lcq}{LC-QuAD 2.0 }

\usepackage[T1]{fontenc}

\usepackage[utf8]{inputenc}

\usepackage{microtype}

\title{Investigating the use of Paraphrase Generation for Question Reformulation in the FRANK QA System}

\author{Nick Ferguson, Liane Guillou, Kwabena Nuamah \and Alan Bundy \\ School of Informatics, University of Edinburgh \\ 10 Crichton Street, Edinburgh EH8 9AB \\ \texttt{nick.ferguson@ed.ac.uk \space liane.guillou@ed.ac.uk} \\ \texttt{k.nuamah@ed.ac.uk \space a.bundy@ed.ac.uk}}

\begin{document}
\maketitle
\begin{abstract}
We present a study into the ability of paraphrase generation methods to increase the variety of natural language questions that the FRANK Question Answering system can answer. We first evaluate paraphrase generation methods on the \lcq dataset using both automatic metrics and human judgement, and discuss their correlation. Error analysis on the dataset is also performed using both automatic and manual approaches, and we discuss how paraphrase generation and evaluation is affected by data points which contain error. We then simulate an implementation of the best performing paraphrase generation method (an English-French backtranslation) into FRANK in order to test our original hypothesis, using a small challenge dataset. Our two main conclusions are that cleaning of \lcq is required as the errors present can affect evaluation; and that, due to limitations of FRANK's parser, paraphrase generation is not a method which we can rely on to improve the variety of natural language questions that FRANK can answer.
\end{abstract}

\section{Introduction}
Within NLP, paraphrasing is the task of generating an output which captures the semantics of an input sentence or question, but with different lexical and/or syntactic features. One significant application of paraphrase generation is Question Answering (QA) (\citet{fader-etal-2013-paraphrase}, \citet{fader14open}, \citet{berant-liang-2014-semantic}, \citet{dong-etal-2017-learning}). The rationale behind the use of paraphrasing is that a user may express a given question in multiple ways, but potentially not in a form that some particular QA system can answer. Paraphrase generation is therefore employed within these systems to improve their ability to handle a wider variety of natural language questions.

The FRANK (Functional Reasoner for Acquiring New Knowledge) QA system combines deductive and statistical reasoning algorithms to produce answers to users' questions when no direct knowledge lookup is available \cite{nuamah2020explainable}. For example, consider the question \textit{What will be the population of France in 2028?}. Clearly, a direct lookup for this piece of information will not be possible\footnote{At the time of writing.}, so, in an effort to provide the user with an answer, FRANK will look up past population data and apply a regression over it.

FRANK is somewhat limited by its natural language capabilities in that it relies on questions to be in a specific form in order to be successfully answered. The above question may be paraphrased as \textit{How many people will be living in France in 2028?} - a formulation which targets the same piece of knowledge, but using a different surface form. However, FRANK's parser will not be able to parse the question into its internal representation, and therefore not be able to generate an answer.

Motivated by previous success in using paraphrase generation for QA tasks, we propose the use of paraphrase generation to transform input questions which FRANK cannot parse due to their form, such as the one above, into forms which FRANK's parser \textit{can} parse. We explore the strengths and weaknesses of selected paraphrase generation methods: backtranslation via Neural Machine Translation (NMT) models; the \textsc{Separator} model of \citet{hosking-lapata-2021-factorising}; and the Paraphrase Database (PPDB) 2.0 \cite{pavlick-etal-2015-adding}. More detail about these methods is given in section \ref{sec:para-gen}. The use of existing methods, as opposed to the creation of a new one, shifts the focus of this study to an analysis of each method's ability to solve the problem of FRANK not being able to answer questions due to their form.

The main contributions of this study are:

\begin{itemize}
    \item Evaluation and comparison of paraphrase generation methods on the \lcq dataset \cite{dubey2019LC} using both automatic and human evaluation.
    \item Error analysis of the \lcq dataset via automatic methods and annotation tasks; and how errors impact paraphrase generation.
    \item Discussion of paraphrase evaluation metrics and their limitations.
    \item Simulation of an implementation of paraphrase generation methods into FRANK and evaluation of their performance on a QA task.
\end{itemize}

\section{Background}
\subsection{FRANK}
\label{subsec:frank-intro}
In FRANK, natural language questions are parsed into a set of attribute-value pairs called an \textit{association list}, or \textit{alist}. FRANK currently uses a template-based parser, but a neural approach following \citet{dong-lapata-2018-coarse} is being integrated \cite{yanzhoudiss}. While this new approach does improve question-alist translation, the question forms that it can parse are still limited, and rely on use of specific terms (e.g., \textit{population}) which can be looked up in knowledge bases.

Attributes in alists include the familiar subject ($s$), property ($p$) and object ($o$) triple, as well as a value attribute, $v$, which can hold the result of a knowledge base query, or the output of an operation (e.g., sum, average, or regression) as described by an attribute $h$. The time attribute, $t$, holds the time period at which to access the property. The alist for the example question \textit{What will the population of France be in 2028?} is given in example \ref{eq:intro_alist}. The inference function $h$ simply returns the population value \texttt{?y0}, which, in this case, is the same as the object.

\begin{equation}
\label{eq:intro_alist}
\begin{split}
\mathcal{A} = \{& \langle \texttt{s, "France"} \rangle,\\
 &   \langle \texttt{p, "population"} \rangle, \\
 &   \langle \texttt{o, ?y0} \rangle, \\
 &   \langle \texttt{t, 2028} \rangle, \\
 &   \langle \texttt{h, "value"} \rangle, \\
 &   \langle \texttt{v, ?y0} \rangle\} \\
\end{split}
\end{equation}

To answer a question with FRANK, alists are recursively decomposed into an \textit{inference graph }according to a set of rules. The resulting graph's leaf nodes represent direct queries to knowledge bases. Values resulting from these queries are aggregated, again according to a set of rules, and then propagated back up the graph to form an answer.


\begin{table*}
    \small
    \centering
    \begin{tabular}{p{0.15\textwidth} p{0.55\textwidth} p{0.2\textwidth}}
    \toprule
        \textbf{Question type} & \textbf{Example} & \textbf{\lcq} \textbf{type} \\
    \midrule
        Simple & \textit{What is the GDP of Ethiopia?} & Single-fact \\
        Nested & \textit{Was the population of France in 2012 greater than the population of Germany in 2009?} & Two-intention \\
        Ranking & \textit{Which country in Africa has the lowest urban population?} & Ranking \\
        Counting & \textit{How many countries border Mexico?} & Count \\
        Boolean & \textit{Did Australia's GDP exceed £400 in 2012?} & Boolean \\
    \bottomrule
    \end{tabular}
    \caption{Types of question that FRANK can answer, with corresponding names in the \lcq dataset.}
    \label{tab:question-types}
\end{table*}

FRANK can answer different types of questions. Such types include simple, single-fact questions; nested comparisons requiring sub-goals to be solved before answering the main question; ranking questions requiring the highest or lowest value to be returned; boolean questions requiring a True or False answer; and counting questions requiring enumeration. In section \ref{subsec:paraphrasing-datasets}, we introduce the \lcq dataset \cite{dubey2019LC} which we use for paraphrase evaluation as it contains question types similar to those that FRANK can answer. Examples of these question types, along with the name of that type in \lcq, are given in table \ref{tab:question-types}.

\subsection{Paraphrasing for Question Answering}

Paraphrase generation has been implemented into QA systems in a variety of ways. \citet{fader-etal-2013-paraphrase} utilise a corpus of paraphrase clusters in order to induce lexical equivalences for relations (e.g., \emph{authored by}, \emph{written by}) and entities (e.g., \emph{President Obama}, \emph{Barack Obama}). Hand-crafted question templates are then used to create paraphrases. Following this template-based approach to paraphrase generation, and using the same paraphrase corpus, \citet{fader14open} mine paraphrase operators in order to induce syntactic variation to question forms. \citet{berant-liang-2014-semantic} generate intermediate logical forms from an input utterance, from which candidate paraphrases are then generated and scored using models trained on question-answer pairs.

QA systems can also take the form of neural models, which are trained on question-answer pairs \cite{dong-etal-2015-question}. In addition to this general approach, \citet{dong-etal-2017-learning} learn paraphrases by implementing a scoring method to predict their quality.

\subsection{Paraphrase generation methods}
\label{sec:para-gen}

Paraphrase identification and generation methods have traditionally used unsupervised, rule-based pattern matching algorithms (\citet{lin_pantel_2001}, \citet{barzilay-mckeown-2001-extracting}, \citet{barzilay-lee-2003-learning}). Recent paraphrase generation methods are based on neural architectures, such as NMT-based backtranslation (\citet{mallinson2017paraphrasing}, \citet{wieting-etal-2017-learning}) and \textsc{Separator} \cite{hosking-lapata-2021-factorising}, both of which are evaluated in this study.

While the template-based methods of \citet{fader-etal-2013-paraphrase} have been shown to be effective, designing templates is labour-intensive and limits the ability of a system to generalise to all possible ways in which a question may be phrased. The very reason for investigating the use of paraphrasing is to free the user from having to ask questions in one of a limited number of specific forms, so while paraphrasing via templates may increase the number of available forms, this method does not allow for the full variety of natural language that users can use to express the same question. Total reformulation of a question via paraphrasing, (e.g., paraphrasing \textit{What is the population of...?} to \textit{How many people live in... ?}), may not be possible using templates unless mappings between forms are created for every possible concept.

\subsubsection{Paraphrasing via Machine Translation}
The introduction of bilingual pivoting \cite{bannard2005paraphrasing} popularised the use of Machine Translation (MT) for paraphrase generation via backtranslation. While originally using Statistical Machine Translation techniques, more modern implementations using NMT models have been explored (\citet{mallinson2017paraphrasing}, \citet{wieting-etal-2017-learning}). Two phrases $p_1$ and $p_2$ are considered paraphrases if both translate to the same phrase in another language, $t$. The NMT models used in this study are those of \citet{OPUS-MT} and are taken from the Huggingface platform \cite{wolf2019huggingface}.

\subsubsection{\textsc{Separator}}
\citet{hosking-lapata-2021-factorising} aim to generate paraphrases of a different surface form from a given input using separate encoding spaces for semantics and syntax. It is this form-changing property which makes \textsc{Separator} a good match for the task of paraphrasing questions in FRANK. With this method, from an input question \textit{How much do football players earn?}, a paraphrase \textit{What do football players get paid?} can be generated.


\subsubsection{The Paraphrase Database(s)}
Moving from paraphrasing whole sentences to paraphrasing single- and multi-word phrases, the Paraphrase Database (PPDB) \cite{ganitkevitch-etal-2013-ppdb} contains millions of lexical, phrasal, and syntactic relations extracted from bilingual parallel corpora using bilingual pivoting. Building on this, in the PPDB 2.0 \cite{pavlick-etal-2015-adding} semantics are introduced in the form of relationships between paraphrase pairs $\langle p_1,p_2 \rangle$, e.g., equivalence ($p_1 \equiv p_2$), hyponymy ($p_1 \sqsubset p_2)$, hypernymy ($p_1 \sqsupset p_2)$ and antonymy ($p_1 =\lnot p_2)$.

\section{Evaluation on \lcq}
\subsection{Paraphrasing datasets}
\label{subsec:paraphrasing-datasets}
Prior to integration into FRANK, paraphrase generation methods described in section \ref{sec:para-gen} need to be evaluated against references. As FRANK can answer different types of question, a requirement of any paraphrase generation method is that it is proficient in paraphrasing these different types, which differ in length and complexity. Therefore, a requirement of any dataset over which paraphrase generation methods are evaluated is that it contains questions in such a variety.

While a variety of paraphrasing-oriented datasets exist, such as the Microsoft Research Paraphrase Corpus \cite{dolan-brockett-2005-automatically}; WebQuestions \cite{fader-etal-2013-paraphrase}; ComQA \cite{abujabal-etal-2019-comqa}; \textsc{GraphQuestions} \cite{su2016generating} and the Quora Question Pairs dataset\footnote{\texttt{https://quoradata.quora.com/First\\-Quora-Dataset-Release-Question-Pairs}}, these do not explicitly distinguish between question types. We instead evaluate paraphrase generation methods on the \lcq dataset \cite{dubey2019LC}, contains over 30,000 data points split between 10 question types, a subset of which correspond to question types that FRANK can answer (see table \ref{tab:question-types}).

Each data point in \lcq contains a natural language question (the \textit{source question}); a paraphrase of that question (the \textit{reference paraphrase}); SPARQL\footnote{\texttt{https://www.w3.org/TR/sparql11-query/}} queries for answering the question using Wikidata \cite{wikidata} and DBpedia \cite{lehmann2015dbpedia}; and metadata describing the question's type. Question types include simple questions (e.g., \emph{What is a red blood cell's mean lifetime?}), and more complex ones with multiple intentions (e.g., \emph{What inspired the creator of the Statue of Liberty, and who is its current owner?}). Table \ref{tab:lcq_amt} shows the process of question and paraphrase generation: SPARQL queries ($Q_S$) were first translated in to hybrid natural language/template forms ($Q_T$) using rule-based methods. Amazon Mechanical Turk (AMT) workers then created natural language verbalisations ($Q_V$) of these hybrid forms; and then paraphrases ($Q_P$) of the natural language verbalisations.

\begin{table}
    \centering
    \begin{tabular}{p{0.12\columnwidth} p{0.73\columnwidth}}
    \toprule
        \textbf{Stage} & \textbf{Example} \\
    \midrule
        $Q_S$ & \texttt{select distinct ?answer where \{ wd:Q37187 wdt:P2645 ?answer\}}\\
    \midrule
        $Q_T$ & What is \texttt{\{mean lifetime\}} of \texttt{\{red blood cell\}} ?\\
    \midrule
        $Q_V$ & Which is the mean lifetime for red blood cell? \\
    \midrule
        $Q_P$ & What is a red blood cell's mean lifetime? \\
    \bottomrule
    \end{tabular}
    \caption{Process of generating natural language questions from SPARQL queries in the \lcq dataset.}
    \label{tab:lcq_amt}
\end{table}

\subsection{Evaluation metrics}
\label{subsec:eval-metrics}

The size of \lcq requires that automatic metrics be used to evaluate model-generated (\textit{candidate}) paraphrases. The output of an ideal metric should reflect the degree to which the semantics of a candidate paraphrase is different that of the reference, while not penalising lexical and syntactic differences. Again, we choose to analyse two existing metrics (iBLEU and cosine similarity via word embeddings) rather than create a new one.

The BLEU metric \cite{papineni-etal-2002-bleu} has been a standard metric for evaluation of machine translation systems since its inception, and has seen use in the field of paraphrase evaluation (\citet{finch-etal-2005-using}, \citet{wan-etal-2006-using}, \citet{madnani-etal-2012-examining}). Standard BLEU evaluates the candidate translation by measuring the $n$-gram overlap between it and a reference translation. \citet{sun-zhou-2012-joint} propose iBLEU, a paraphrase-oriented modification of BLEU. It aims to create a trade-off between the semantic equivalence of the candidate paraphrase $c$ and reference paraphrase $r$; and the similarity between the candidate and the source $s$. The trade-off is moderated by the constant $\alpha=0.7$, a value derived from human evaluation. The formula is shown in equation \ref{eq:ibleu}.
-
\begin{equation}
\begin{split}
    \textrm{iBLEU} = & \alpha\textrm{BLEU}(c, r) \\
     & - (1-\alpha)\textrm{BLEU}(c, s)
\end{split}
\label{eq:ibleu}
\end{equation}

With the acknowledgement that some perfectly valid paraphrases may bear little-to-no resemblance to the source phrase or a reference paraphrase from an $n$-gram overlap point of view, evaluation using word embeddings was employed in an effort to capture semantic similarity. We use the sentence embeddings of \citet{reimers-gurevych-2019-sentence} to compute cosine similarity between candidate paraphrases and source questions. We also analyse the metrics themselves by computing correlation between the metrics and human judgement. This is to confirm that they are equally good at evaluating paraphrases of simple, short questions as well as longer, more complex ones.

The choice of these evaluation metrics allow us to approach evaluation from a syntactic and semantic perspective, with the aim of building a more comprehensive picture of the performance of each paraphrase generation method. These open-source metrics have applications across multiple tasks, which help improve their interpretability. An alternative could be ParaMetric \cite{callison-burch-etal-2008-parametric} which is based off of word alignment. However, this may not be a good measure of performance considering that we desire a high degree of form-changing from a paraphrasing method, which ParaMetric may not be able to evaluate.

\subsection{Experiment}

We generated paraphrases using the methods described in section \ref{sec:para-gen} from the source questions in \lcq, then evaluated on the reference paraphrases from \lcq using metrics described in \ref{subsec:eval-metrics}. Rather than selecting a single language for the backtranslation method at the outset, we tested different languages from different families: German (\textsc{de}), French (\textsc{fr}), Hindi (\textsc{hi}), Russian (\textsc{ru}), and Chinese (\textsc{zh}). The source question from a given data point was translated from English (\textsc{en}) into one of the target languages, then translated from the target language back to English. 

Firstly, we split into different question types using metadata from each data point (as described in section \ref{subsec:paraphrasing-datasets}). We found no direct correspondence between the 10 question types described in \citet{dubey2019LC} and the set of values present for each of the metadata attributes; so we manually divided the dataset into 7 question types, 5 of which are types that FRANK can answer: \textit{single-fact}, \textit{two-intention}, \textit{ranking}, \textit{boolean}, and \textit{counting} questions (as described in section \ref{subsec:frank-intro}). We focused on these question types in our experiments. While other question types not analysed here also correspond to the \textit{nested} type in table \ref{tab:question-types}, we aligned this type to the \textit{two-intention} type of \lcq as it presented longer, more complex questions than other types in the dataset.

\subsection{Results}
\label{subsec:lcq-results}

\subsubsection{Automatic evaluation}
\label{subsubsec:lcq-auto-eval}
Average cosine similarity between each candidate paraphrase and source sentence is shown, for each question type, in table \ref{tab:cossim_all_models_all_types}. The best model for each type and a weighted average is also shown. Despite differences in overall performance for each model, performance across question types was consistent to within 3 percentage points of the average for the best performing model.

For backtranslation-based paraphrasing, it is likely that training data availability, as well as language similarity to English, influenced candidate paraphrase quality. Of the language pairs tested here, those which are most similar to English (French and German) performed better on both cosine similarity and human judgement. The \textsc{en-ru} and \textsc{en-zh} pairs performed similarly, both approximately 10\% worse than the \textsc{en-fr} and \textsc{en-de} models. The \textsc{en-hi} performed significantly worse, with human evaluation (detailed in section \ref{subsubsec:lcq-human-eval}) showing that it rarely produced adequate output. These poorer-performing language pairs are further in similarity from English in terms of morphological richness and strictness of word ordering, when compared to French or German. In addition to linguistic similarity, the quantity of training data likely influenced translation quality. For example, only 26.9M sentence pairs were available for \textsc{en-hi}, in contrast to the 479M for \textsc{en-fr}.

\textsc{Separator} performed much worse than the top-performing backtranslation methods. Human annotation showed that backtranslation methods frequently produce paraphrases identical to source questions, while \textsc{Separator} produced very few, preferring to produce poor paraphrases. This is a result of the training objective of each model: translation models are trained to produce an output that is as close to the semantics of the input, while ensure the syntax of the output is grammatical. However, \textsc{Separator} is specifically trained to alter syntax while preserving semantics. Results on the cosine similarity metric show that semantics are not preserved as well as backtranslation-based paraphrasing.

\begin{table*}
    \small
    \centering
    \begin{tabular}{ccccccc}
    \toprule
        \textbf{Model} & \textbf{Boolean} & \textbf{Count} & \textbf{Rank} & \textbf{Single-fact} & \textbf{Two-intent.} & \textbf{Average}\\
    \midrule
        \textsc{en-de} & 0.87 $\pm$ 0.12 & \textbf{0.91} $\pm$ \textbf{0.10} & 0.85 $\pm$ 0.13 & \textbf{0.86} $\pm$ \textbf{0.14} & \textbf{0.88} $\pm$ \textbf{0.11} & \textbf{0.88} $\pm$ \textbf{0.13} \\
        \textsc{en-fr} & \textbf{0.87} $\pm$ \textbf{0.12} & 0.90 $\pm$ 0.10 & \textbf{0.86} $\pm$ \textbf{0.14} & 0.86 $\pm$ 0.14 & 0.88 $\pm$ 0.12 & 0.88 $\pm$ 0.13\\
        \textsc{en-hi} & 0.52 $\pm$ 0.22 & 0.53 $\pm$ 0.22 & 0.56 $\pm$ 0.23 & 0.53 $\pm$ 0.24 & 0.53 $\pm$ 0.22 & 0.53 $\pm$ 0.23\\
        \textsc{en-ru} & 0.80 $\pm$ 0.16 & 0.84 $\pm$ 0.14 & 0.80 $\pm$ 0.17 & 0.78 $\pm$ 0.17 & 0.83 $\pm$ 0.14 & 0.81 $\pm$ 0.16\\
        \textsc{en-zh} & 0.77 $\pm$ 0.17 & 0.82 $\pm$ 0.15 & 0.76 $\pm$ 0.18 & 0.76 $\pm$ 0.18 & 0.79 $\pm$ 0.16 & 0.77 $\pm$ 0.17\\
        \textsc{Separator} & 0.59 $\pm$ 0.19 & 0.66 $\pm$ 0.17 & 0.62 $\pm$ 0.18 & 0.66 $\pm$ 0.19 & 0.72 $\pm$ 0.17 & 0.65 $\pm$ 0.18\\
    \bottomrule
    \end{tabular}
    \caption{Average cosine similarities between source questions and candidate paraphrases generated by each model, over different question types, $\pm$ one standard deviation. The best model for each question type is shown in bold.}
    \label{tab:cossim_all_models_all_types}
\end{table*}

\subsubsection{Human evaluation}
\label{subsubsec:lcq-human-eval}
\begin{figure}[!ht]
    \centering
    \includegraphics[width=\columnwidth]{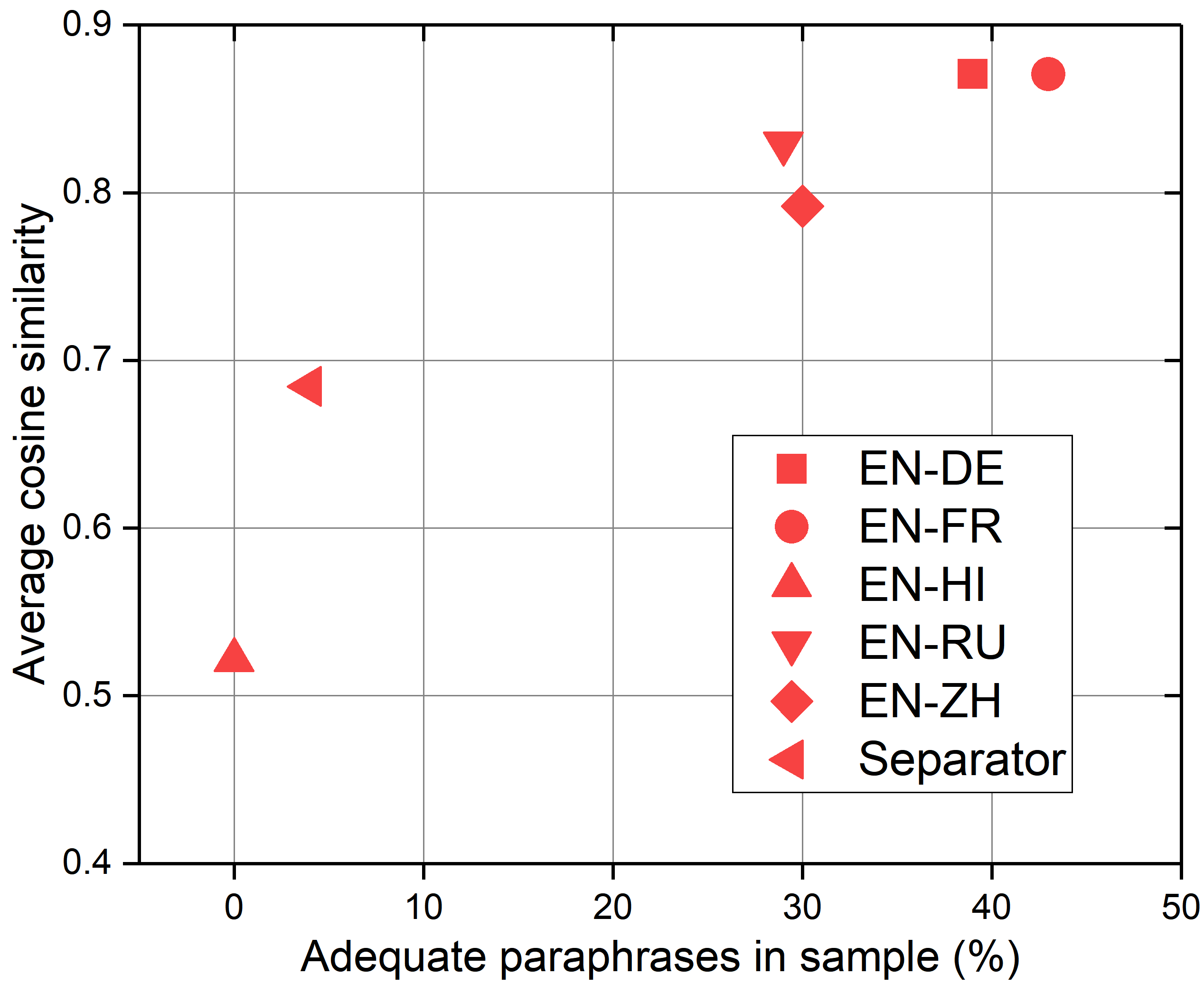}
    \caption{Correlation between human and automatic evaluation for model-generated paraphrases from \textit{two-intention} questions.}
    \label{fig:adeq-cossim}
\end{figure}

To confirm that cosine similarity and iBLEU accurately assess the quality of generated \textit{questions} of different complexities (as opposed to \textit{sentences}), we assessed the adequacy of candidate paraphrases via manual annotation. We randomly sampled a set of 100 questions from both the question types \textit{single-fact} and \textit{two-intention}: representing both shorter and simpler questions; and longer, more complex questions respectively. We generated candidate paraphrases using each model in table \ref{tab:cossim_all_models_all_types}, then annotated them as being either \textit{adequate}, \textit{inadequate}, or \textit{trivial}. Similar to the definition of adequacy in MT evaluation, a paraphrase is \textit{adequate} if it captures the intention of the source question from which it was generated. An \textit{inadequate} paraphrase therefore does not capture the intention of the source question. Different degrees of inadequacy exist. It may be caused, for example, by a lack of fluency in an otherwise adequate paraphrase, or a severe loss of information rendering it unrecognisable with respect to the source question. A \textit{trivial} paraphrase is either identical to the source question, or, for example, contains trivial word replacements such as \textit{Who's} $\rightarrow$ \textit{Who is}. Figure \ref{fig:adeq-cossim} shows the correlation between adequacy and cosine similarity for paraphrases generated from the \textit{two-intention} set. After candidate paraphrases annotated as trivial were removed from the sets, the cosine similarity between candidate paraphrases and source questions was computed and averaged, forming the \textit{y}-axis in figure \ref{fig:adeq-cossim}. The \textit{x}-axis is the number of adequate paraphrases that were generated by each model, expressed as a percentage. Spearman's $\rho$ is shown for human and automatic evaluation, for both question sets and both automatic metrics, in table \ref{tab:human-auto-eval-spearman}. Figure \ref{fig:adeq-cossim} therefore corresponds to the Spearman's $\rho$ in the \textit{two-intention}, \textit{cosine similarity} cell. Equivalent figures for the remaining cells are given in appendix \ref{app:restofthegraphs}.

\begin{table}[!ht]
    \centering
    \begin{tabular}{ccc}
    \toprule
        \textbf{Question type} & \textbf{Cosine similarity} & \textbf{iBLEU} \\
        \midrule
        Single-fact & 0.812 & 0.464 \\
        Two-intention & 0.943 & 0.029 \\
    \bottomrule
    \end{tabular}
    \caption{Spearman's $\rho$ between the frequency of adequate paraphrases and each of the automatic metrics, for two question types.}
    \label{tab:human-auto-eval-spearman}
\end{table}

\begin{figure}[!ht]
    \centering
    \includegraphics[width=\columnwidth]{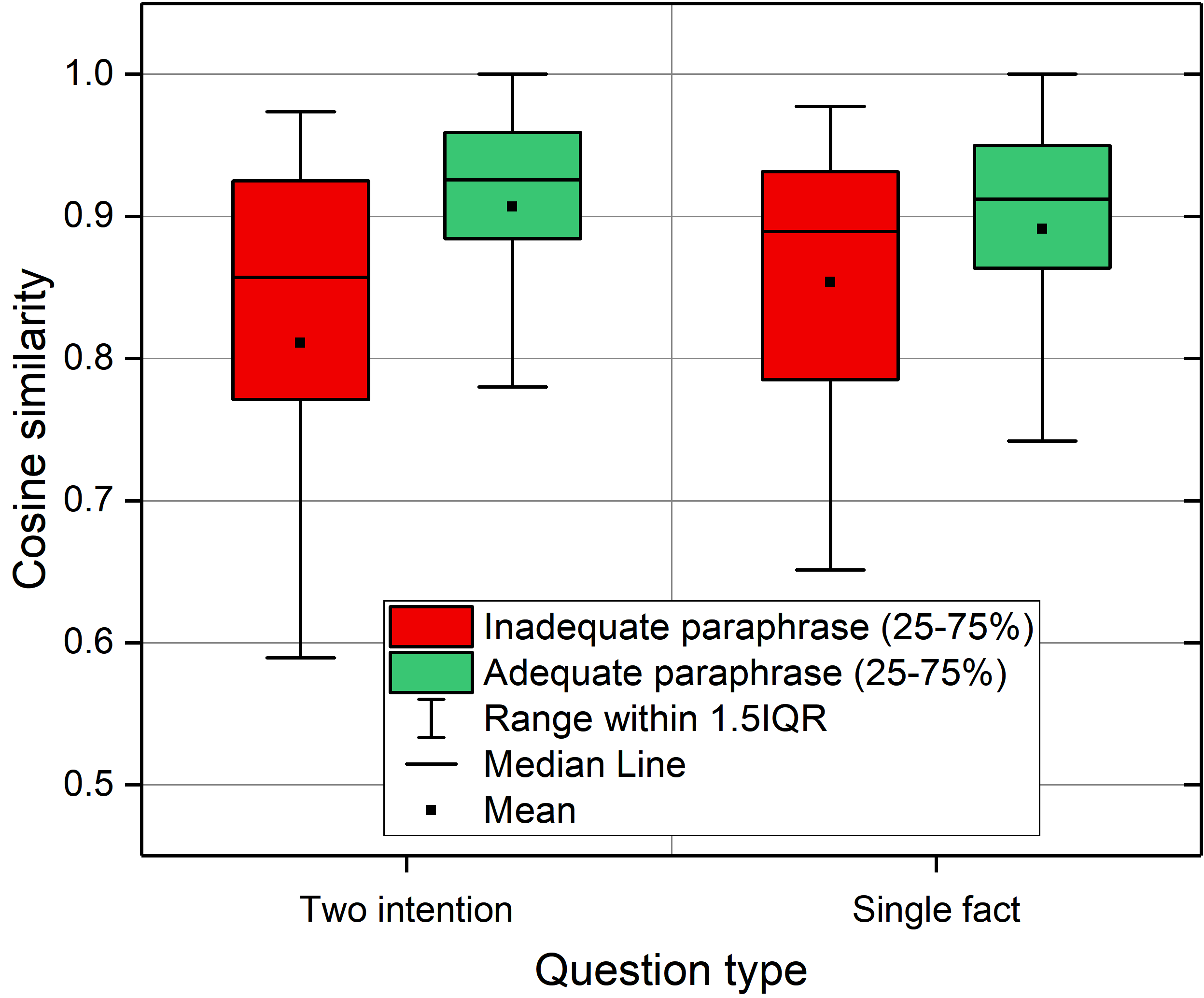}
    \caption{Distribution of cosine similarities for adequate and inadequate paraphrases over both question types.}
    \label{fig:adeq-inad-boxplots-overlap}
\end{figure}

For both question types, table \ref{tab:human-auto-eval-spearman} shows a strong correlation between the frequency of adequate paraphrases generated by a model and the average cosine similarity between candidate paraphrases and source questions. While this correlation does suggest that models whose paraphrases have high cosine similarity to source questions are more likely to be adequate; even models the highest cosine similarities only produce adequate paraphrases at a frequency of less than 50\%. As averaging cosine similarities across adequate and inadequate paraphrases in figure \ref{fig:adeq-cossim} is a crude operation to perform, figure \ref{fig:adeq-inad-boxplots-overlap} highlights the distribution of cosine similarities for the adequate and inadequate sets for both question types. While adequate paraphrases do have higher cosine similarities, than the inadequate paraphrases, there is a high degree of overlap. We expect the reason for this to be that, for sentence embeddings, semantic contribution appears to be greater for non-stopwords \cite{ethayarajh-2019-contextual}. This is despite the fact that stopwords often contain the querying words, (e.g., \emph{what}, \emph{which}, \emph{how}). The intention of two questions, or the quantities which they target, can differ dramatically by the replacement these words, while the coarse-grained objective remains the same. For example, the cosine similarity between the questions \emph{`What is the Statue of Liberty?'} and \emph{`Where is the Statue of Liberty?'} is 0.93. The questions target distinctly different pieces of information, but they are similar to the extent that both are asking a question about the same subject.

Correlation between iBLEU and frequency of adequate paraphrases was extremely poor, meaning that a high iBLEU score is not indicative of an adequate paraphrase. While correlation between cosine similarity and human judgement was high for both question types, this was not the case for iBLEU. Length and complexity of the \textit{two-intention} questions could be a factor in its exceptionally poor correlation. As for the overall poor performance, one reason for this may be that we are limited to using iBLEU in a single-reference setting. The original BLEU metric was designed to be computed against multiple reference translations, in order to account for the many valid translations for a given sentence. However, \citet{freitag-etal-2020-bleu} state that high BLEU scores do not correlate with higher quality translations and suggest that the nature of the references are more important than their number. This supports a second reason, which is that reference paraphrases of a given source question may be either syntactic, lexical/phrasal, or both. A candidate syntactic paraphrase will achieve low iBLEU score when computed against a reference phrasal paraphrase. For example, consider a source question \textit{What is the population of France?}. A candidate syntactic paraphrase \textit{How many people live in France?} will evaluate poorly if compared against a reference phrasal paraphrase \textit{What is the total population of France?}, despite the candidate paraphrase being perfectly adequate.

\subsection{Error analysis}

During examination of a number of data points within \lcq, it became apparent that a number of issues exist in the dataset. Some errors, such as empty data fields, can be detected through automatic methods (summarised in table \ref{tab:lcq-auto-error}); while others, such as ungrammatical paraphrases, require a manual approach to ensure detection with as high an accuracy as possible (summarised in table \ref{tab:lcq-annotation}).

In anticipation that these problematic data points may affect automatic evaluation, we performed error analysis in the form of an annotation study, described in section \ref{sec:lcq-annotation}.

\subsubsection{Automatic methods}
\label{sec:auto_error}

Some error types were only present in a small number of items, such as the 0.1\% of data points which contain file extensions (e.g., \texttt{.jpg}) in either the question or the paraphrase; and the 0.5\% of data points that have an empty field (or some variation of \texttt{N/A}) in place of the question or the paraphrase.

In 1\% of the data points, characters with accents, e.g., \textit{\"{u}} or \textit{\'{e}} were not reproduced in the paraphrase, but instead replaced with the basic letterforms, e.g., \textit{u} or \textit{e}. The consequence of this is that named entities, such as \textit{L\"{u}beck} or \textit{Jos\'{e}} are not identically reproduced in the paraphrase, though they should be.

The dataset contained errors resulting from the translation task shown in table \ref{tab:lcq_amt}: some natural language questions, $Q_V$, and their paraphrases, $Q_P$, still contained these template-like terms from  $Q_T$ which have curly braces (e.g., \texttt{\{red blood cell\}}). 5.8\% of data points had these template-like terms in what are supposed to be natural language questions and/or paraphrases.

6.8\% of data points contain a paraphrase which is identical to the question, which mean that any model-generated paraphrases evaluated using iBLEU are not evaluated in a way intended by the metric. iBLEU scores are therefore artificially high for any candidate paraphrase generated from and evaluated on these data points.

In total, 13.5\% of data points can be rejected due to errors discovered by these automatic methods. The rationale for rejecting these data points is that they may present an issue for either the generation of paraphrases from the questions, or the evaluation of those generated paraphrases with respect to the reference paraphrase - a hypothesis which is explored in section \ref{sec:error-paraphrases}.

\begin{table}
    \centering
    \begin{tabular}{cc}
        \toprule
        \textbf{Error category} & \textbf{Frequency} (\%) \\
        \midrule
        File extensions & 0.1 \\
        Empty field & 0.5 \\
        Missing accents & 1.0 \\
        Template-like terms & 5.8 \\
        Identical paraphrase & 6.8 \\
        \midrule
        \textbf{Total} & \textbf{13.5} \\
        \bottomrule
    \end{tabular}
    \caption{Distribution of errors found using automatic methods. 13.5\% of all data points contained errors in some form which could be automatically extracted.}
    \label{tab:lcq-auto-error}
\end{table}

\subsubsection{Manual error analysis}
\label{sec:lcq-annotation}

To expand on the automatic error analysis performed in section \ref{sec:auto_error}, further analysis was conducted in the form of an annotation task. The goal of this task was to isolate four sets of data points based on the quality of the question and its reference paraphrase in order to assess the prevalence of error which could not be determined by automatic means, e.g., spelling mistakes. One set contains data points which contain \textit{no error}, that is, the paraphrase perfectly captures the intention of the question; both questions are grammatical; named entities are reproduced correctly; and there are no spelling mistakes. The remaining three sets contain data points which contain some form of error in the \textit{question}, in the \textit{paraphrase}, or in both \textit{question and paraphrase}. Phenomena which were classed as errors include any error which would have been captured by the automatic methods in section \ref{sec:auto_error}, as well as ungrammaticalities; spelling mistakes; and invalid paraphrasing of named entities or historic events. Examples include \textit{Neil Precious stone} as a paraphrase of \textit{Neil Diamond}, and \textit{the Cuban Rocket Emergency} as a paraphrase of \textit{the Cuban Missile Crisis}. This annotation was performed on 608\footnote{Data was annotated until at least 50 data points were present for each category in table \ref{tab:lcq-annotation}.} data points belonging to the \textit{single-fact} question type and the results are summarised in table \ref{tab:lcq-annotation}. 

\begin{table}
    \centering
    \begin{tabular}{cc}
        \toprule
        \textbf{Error category} & \textbf{Frequency} (\%) \\
        \midrule
        No error & 23.7 \\
        Question & 8.2 \\
        Paraphrase & 47.2 \\
        Question and paraphrase & 20.1 \\
        \bottomrule
    \end{tabular}
    \caption{Distribution of error among questions of the \textit{single-fact} type. 76.3\% of data points contained some form of error in the question or the paraphrase.}
    \label{tab:lcq-annotation}
\end{table}

When training a QA system on question-answer pairs, a certain proportion of ungrammatical questions is somewhat desirable. Users will not always ask perfectly formed, grammatical questions, so it is important that a QA system be robust to this. However, this was but a subset of the questions or paraphrases which contained ungrammatical or poorly formed questions. Paraphrases in particular contained egregious substitutions for words in the source sentence, many of which no native English speaker would reasonably make. Some examples are given in appendix \ref{app:lcq-examples}, which highlight an issue with using AMT for this task in that there is no way of automatically verifying the quality of these human-generated paraphrases.

\subsubsection{Effect of question errors on paraphrasing}
\label{sec:error-paraphrases}

\begin{figure}[!ht]
    \centering
    \includegraphics[width=\columnwidth]{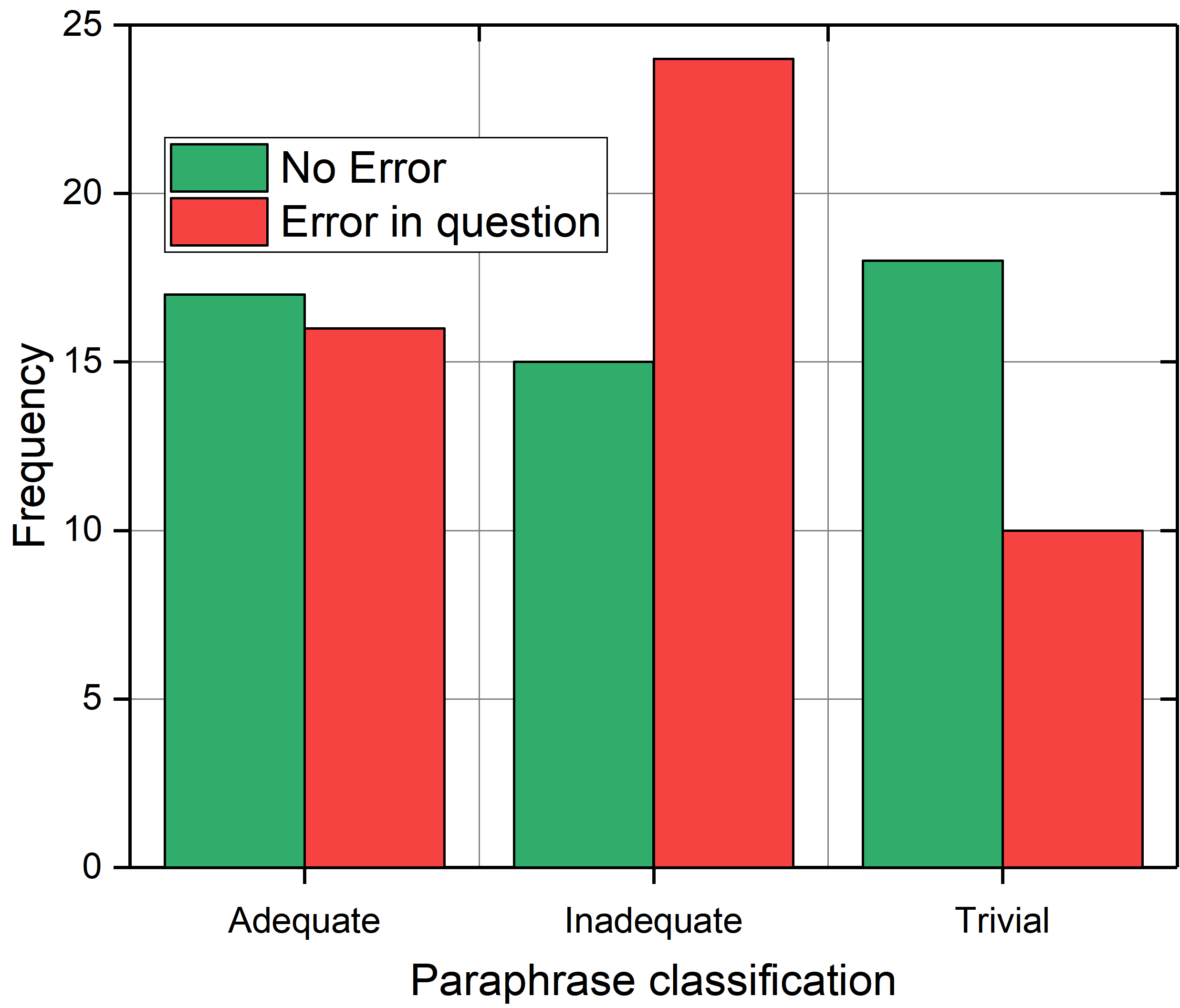}
    \caption{Classification of paraphrases generated using the \textsc{en-fr} model when using source questions containing/not containing error.}
    \label{fig:error-comparison-annotation-results}
\end{figure}

After establishing that 76.3\% of data points from the \textit{single-fact} type contain some form of error in the question or the paraphrase; it must be established whether this impacts the quality of generated paraphrases. As NMT models are trained on grammatical sentence pairs, it may be expected that a backtranslation-based paraphrasing method will be provide reduced performance on ungrammatical source questions. While we have limited results on the human evaluation stage in section \ref{subsubsec:lcq-human-eval}, we chose the \textsc{en-fr} model as a single model to proceed with as it scored highest in both cosine similarity and human judgement. Using 50 source questions which contain no error, and 50 which do contain error, paraphrases were generated and annotated using the same \textit{adequate}, \textit{inadequate}, and \textit{trivial} classification as in section \ref{subsubsec:lcq-human-eval}. We found that, while each set of source questions produced approximately the same number of adequate paraphrases, the rate of inadequate paraphrases was higher for the paraphrases generated from questions containing errors. These results are summarised in figure \ref{fig:error-comparison-annotation-results}.

There are cases of the paraphrasing process correcting the grammar of ungrammatical source questions. Such repairs include the insertion of missing determiners and prepositions, e.g., \textit{What is half-life of technetium-99m?} is paraphrased to \textit{What is \textbf{the} half-life of technetium-99m?}. Classification of candidate paraphrases as inadequate was largely due to ungrammaticalities in the source propagating to the output, rather than any information loss occuring during translation. The relatively high rate of trivial paraphrases for the source questions without error represents a large proportion of paraphrases identical to the source question, or with trivial replacements, e.g., \textit{Which} $\rightarrow$ \textit{What}. This may be due to the fact the the \textsc{en-fr} and \textsc{fr-en} models were trained on the same parallel data. Training each of these models on different data, or using backtranslation between multiple languages (e.g., \textsc{en-fr}, \textsc{fr-de}, \textsc{de-en}) may result in more varied paraphrases. For some ungrammatical or poorly formed source questions, e.g., \textit{who concept for studied by of materials science?}, classifying the candidate paraphrase required inferring what a given source question might mean, possibly leading to error in classification.

\section{Evaluation on FRANK}
\label{sec:frank-eval}
To test if a particular model will bring about improved performance when integrated into FRANK, we simulated integration of a paraphrase generation method. In FRANK, paraphrase generation methods may be implemented in two places. Firstly, the full input question may be paraphrased using the NMT or the \textsc{Separator} methods; and secondly, elements of the root alist, such as the property or subject, may be replaced with lexical or phrasal paraphrases using the PPDB 2.0. PPDB 2.0-based paraphrasing can therefore take place in addition to the two previous methods discussed, functioning as an optional add-on after the alist has been parsed. We explore this functionality in our experiment.

\subsection{Experiment}

As with section \ref{sec:error-paraphrases}, we choose the \textsc{en-fr} model to create paraphrases. We create a very small test dataset which contains a two source questions for each of the simple and nested question types, along with their alists. For each of these four questions, 5 different hand-created paraphrases are created, encoding the same intent as the original question, but introducing different syntax, or synonyms. Each of these paraphrases are then `re-paraphrased' by the \textsc{en-fr} model, then passed to FRANK. If FRANK's parser returns an equivalent alist (that is, one containing the same attribute-value pairs) as that of the original question, then this constitutes a success. Each human-generated paraphrase is engineered such that FRANK cannot parse it, so the task is to `recover' the original question via paraphrasing.

\subsection{Results}

Out of the 20 test cases, only one candidate paraphrase generated by the \textsc{en-fr} model resulted in the source question's alist, and contrasts the success that has previously been achieved when implementing paraphrasing into QA systems. The source of error was not translation quality - all 20 paraphrases were adequate. Rather, for the model-generated paraphrases that FRANK was able to parse into alists, they were faithful enough to the unparsable human-generated paraphrase that the problem remained that FRANK's parser could not parse them into alists because of their form. For example, one test case was the question \emph{`In 2026, how many people will be living in France?'}, for which the model generated the paraphrase \emph{`In 2026, how many people will live in France?'}. The paraphrase has the same form on a coarse-grained level. In addition to highlighting the brittle nature of FRANK's parser, this highlights a problem with paraphrase generation, in we had no way to specify a target form for a paraphrase. So, while all paraphrases were adequate, they were not sufficiently different from the human-generated paraphrase to return the same alist as the test question, and we had no method by which to transform them into forms which FRANK can parse. While this experiment may be repeated for the neural parser being implemented \cite{yanzhoudiss}, it is still not expected that this will allow for the increase in natural language variation that we had hoped to achieve via paraphrasing.

The dataset was too small to be able to draw conclusions about the usefulness of paraphrasing with the PPDB 2.0. For the model-generated paraphrases that FRANK was able to parse alists from, the property ($p$) field was paraphrased by searching PPDB 2.0 for equivalent words or phrases. This produced alists which could serve as alternative representations of the question if decompostition of the original alist failed to result in an answer. However, in our small sample size, paraphrases discovered using the PPDB 2.0 were not always semantically similar to their source due to lack of contextual information.

\section{Discussion}
The error analysis on \lcq has shown that a significant number of data points are affected by one or more types of error. However, the manual evaluation was limited in two ways. Firstly, only a single annotator (the lead author) performed annotation. This may have introduced bias to the classification of error, for example, if certain questions required specific domain knowledge. Increasing the number of annotators, as well as their diversity in terms of domain knowledge and background, should be a requirement for any future annotation tasks. Secondly, only a coarse-grained analysis was performed in this study, but further analysis may be undertaken on a more fine-grained level in order to better understand the prevalence of different types of error. Such fine-grained classifications may include: spelling mistakes; ungrammaticalities; poorly phrasing of questions or paraphrases; incorrect copying of named entities between questions and paraphrases, etc., for both questions and paraphrases. We will then be able to gain a better understanding of the effect of different types of error on the generation of candidate paraphrases.

The other annotation task in this study was a rating of the adequacy of candidate paraphrases generated by models. This annotation task likely suffered due to the same reasons as the one previously described: a single annotator and coarse-grained classification. Potential fine-grained categories may include the type of each paraphrase produced by a given model. As discussed in section \ref{subsubsec:lcq-auto-eval}, the similarity of English and the pivot language used in backtranslation may influence the type of paraphrase generated (e.g., syntactic or lexical/phrasal).

In the datasets discussed, the type of the reference paraphrase was not denoted. Paraphrases ranged from trivial substitution of query words (e.g., \textit{Which} $\rightarrow$ \textit{What}) to total reformulation of questions. As discussed in section \ref{subsubsec:lcq-human-eval}, this likely affects evaluation via automatic metrics. An improvement to any paraphrasing dataset would be the organisation of multiple reference paraphrases under different types. This will allow for more accurate and fine-grained evaluation of candidate paraphrases. 

Regarding paraphrase evaluation, this investigation has also highlighted the poor ability of some automatic metrics to accurately assess the quality of candidate paraphrases, to the point where human judgement became our primary metric. While we may not have not exhausted all off-the-shelf metrics, we still emphasise the need for an automatic metric with attributes described in \ref{subsec:eval-metrics}: it should reward semantic similarity of a candidate paraphrase to the source from which it was generated, and allow for (that is, not penalise) rich syntactic paraphrasing. In the case of iBLEU, there may be a more appropriate value for the parameter $\alpha$ for our task. The value we chose, 0.7, was the lower bound of the recommended range 0.7-0.9, following \citet{hosking-lapata-2021-factorising}. Lower values punish self-similarity more severely, so given that we are seeking to change question forms via paraphrasing, values lower than that of the recommended range may be worth experimenting with. However, the use of iBLEU is still flawed, even if we do change $\alpha$, because the type of the reference paraphrase in \lcq varied between syntactic and lexical and therefore iBLEU results will be inconsistent. 

Similarly, there is no known method to control the form of a candidate paraphrase - the nature of models' outputs are a product of their training. Rather than changing the surface form of an input question via paraphrasing into some unknown, different form (as with \textsc{Separator}), there may be room for a model which is able to paraphrase inputs in a variety of forms into a single, pre-determined form.

Some questions contain phrases which should not be paraphrased, such as named entities, dates, and historical events. These cases were generally handled well by backtranslation methods, but not by \textsc{Separator}. However, these categories are not the only ones which should be `protected' from paraphrasing - with another category being scientific or technical terminology. Another recent update to FRANK allowed asking questions about specific statistical quantities of data, e.g., coefficients of variance - this is a specific piece of mathematical terminology and any paraphrasing applied to it may result in a loss of meaning. Therefore, another branch of future work with respect to paraphrasing may be the ringfencing of domain-specific terms.

Where we would hope that the PPDB 2.0 paraphrasing method might paraphrase \textit{size} in to \textit{surface area}, it instead returned paraphrases such as \textit{file size}. We could make use of terminology in Knowledge Bases, as well as a context which we could obtain from the subject ($s$) attribute of the alist, to measure the likelihood of being interested in \emph{surface area} or \emph{file size} given the subject \emph{France}.

\section{Conclusion}
In this study, we found that paraphrase generation did not bring about improvements to the variety of natural language questions that FRANK can handle. It became apparent that FRANK's current parser remains a bottleneck, limiting the variety of question formulations that a user can ask. We therefore believe that improving FRANK's parser is a more appropriate direction for future work, as opposed to further work on implementing paraphrasing methods.

We have evaluated different paraphrase generation methods on the \lcq dataset using a combination of automatic metrics and human judgement. We found poor correlation between the iBLEU metric and human evaluation and show that, while cosine similarity correlated highly with human judgement, it did not always indicate a high quality output.

We performed an error analysis on the \lcq dataset, and found, using automatic methods, that over 13\% of data points can be shown to contain error. Further error analysis via human annotation suggested that over 76\% of data points contained error in some form in either the question or its reference paraphrase. We recommended further error analysis in order to better understand the prevalence and nature of the errors present.

While the test implementation into FRANK did not produce positive results, there may be a future for the experiment with some fine-tuning of the methodology. We have discussed how backtranslation using different language familties may results in different types of paraphrases. So, while we chose the best-performing model based on our evaluation, it may be the case that a different language produces paraphrases closer to that which we are seeeking. Additionally, the low precision of the PPDB 2.0 method calls requires consideration of context in order to suggest more semantically similar paraphrases.

\bibliography{custom}

\appendix
\section{Correlation between human evaluation and automatic metrics}
\label{app:restofthegraphs}

\begin{figure}[!h]
    \centering
    \includegraphics[width=0.8\columnwidth]{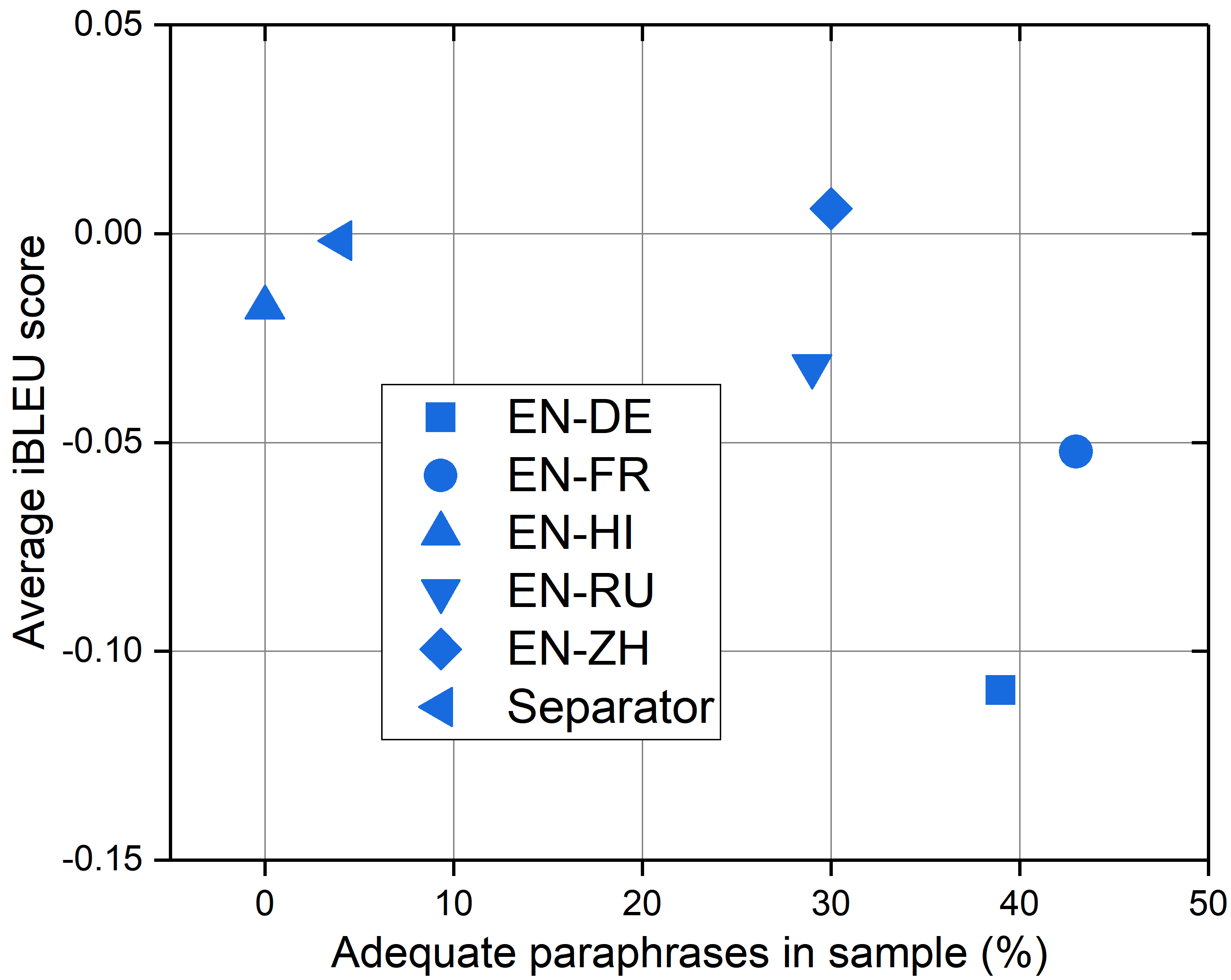}
    \caption{Correlation between human evaluation and iBLEU score for the \textit{two-intention} question type.}
    \label{fig:two-int-ibleu}
\end{figure}

\begin{figure}[!h]
    \centering
    \includegraphics[width=0.8\columnwidth]{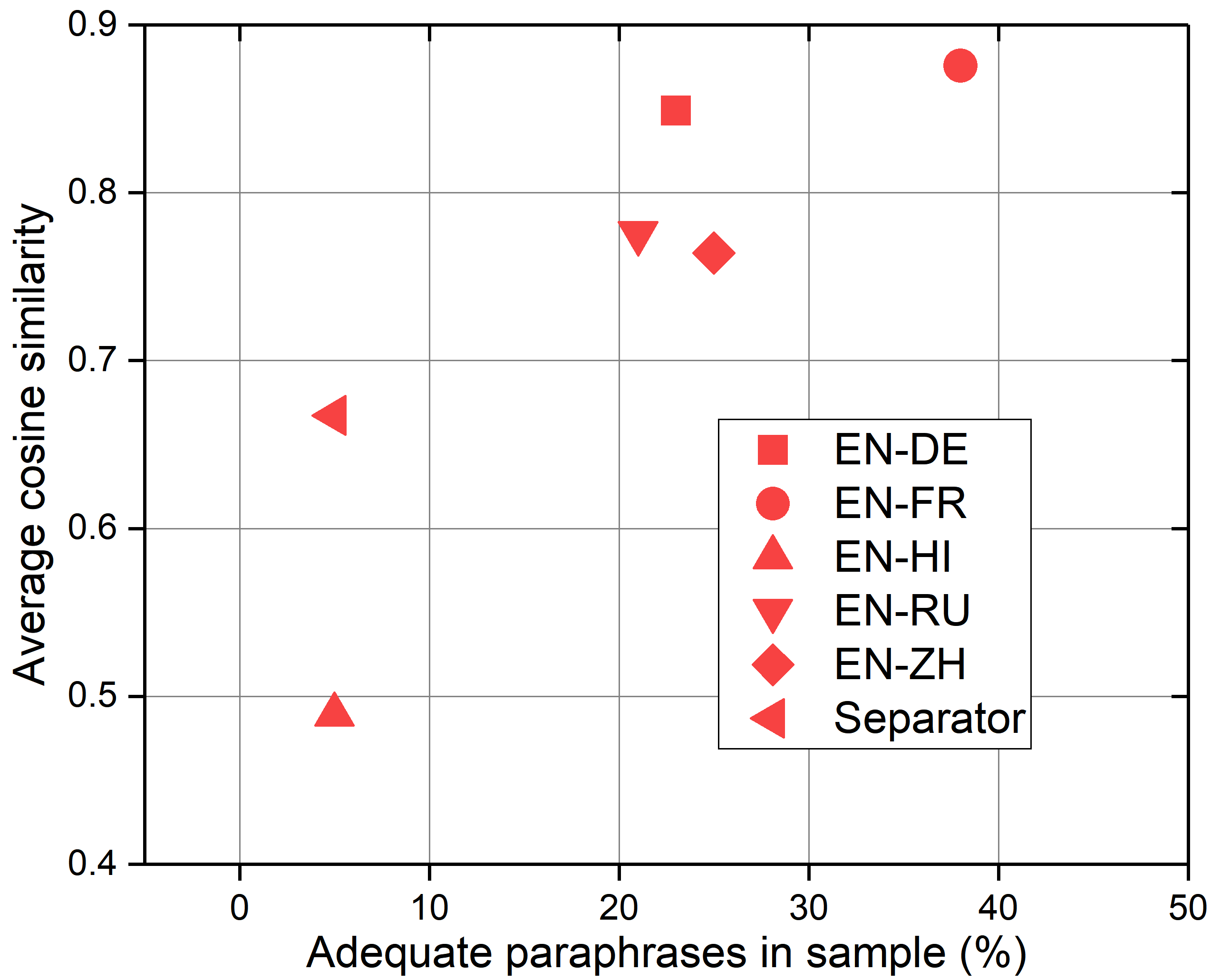}
    \caption{Correlation between human evaluation and cosine similarity for the \textit{single-fact} question type.}
    \label{fig:sin-fac-cossim}

\end{figure}

\begin{figure}[!h]
    \centering
    \includegraphics[width=0.8\columnwidth]{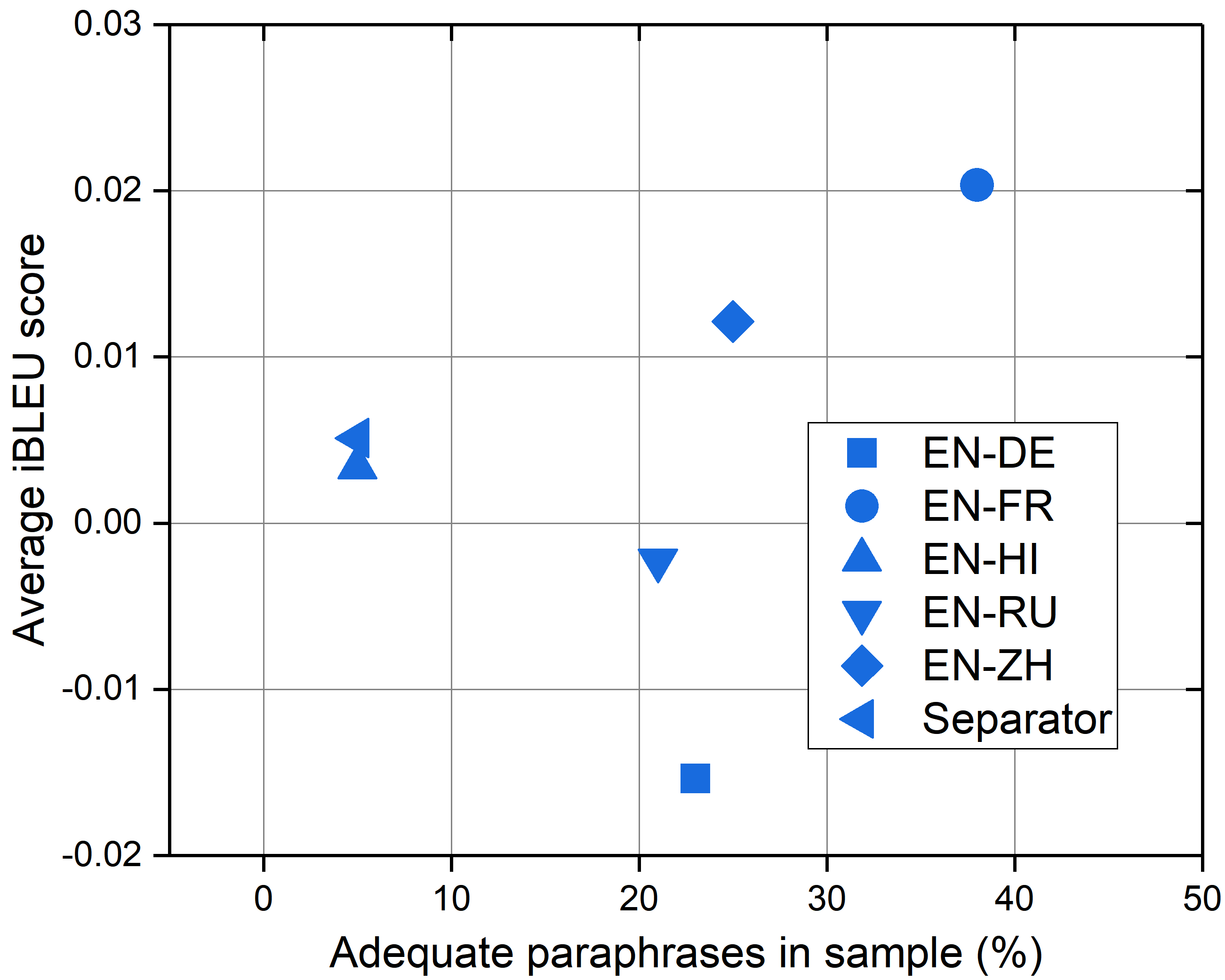}
    \caption{Correlation between human evaluation and iBLEU score for the \textit{single-fact} question type.}
    \label{fig:sin-fac-ibleu}
\end{figure}

Figures \ref{fig:two-int-ibleu}, \ref{fig:sin-fac-cossim} and \ref{fig:sin-fac-ibleu} expand on figure \ref{fig:adeq-cossim} in section \ref{subsubsec:lcq-human-eval} to show the relationship between both metrics and human evaluation, for the \textit{single-fact}  and \textit{two-intention} question types. 

\section{\lcq examples}
\label{app:lcq-examples}

\begin{table*}[!ht]
    \small
    \centering
    \begin{tabular}{p{0.35\textwidth} p{0.35\textwidth} p{0.2\textwidth}}
        \toprule
        \textbf{Question} & \textbf{Reference paraphrase} & \textbf{Error}\\
        \midrule
        What is the cardinality of this set of \textit{rational} numbers? & What is the cardinality of this set of \textit{levelheaded} numbers? & Literal paraphrase \\
        \midrule
        Did \textit{Pliny the Younger} die in Brazil? & Did \textit{Pliny the More youthful} pass on in Brazil? & Named entity \\
        \midrule
        Tell me about award received of \textit{Animal Farm} and followed by? & Let me know approximately grant gotten of \textit{Creature Cultivate} and taken after by? & Literal paraphrase, poorly formed \\
        \midrule
        what is the higher taxon of motif of \textit{fantastic bird} MG 17727 & what is the higher taxon of theme of \textit{phenomenal fowl} MG 17727 & Literal paraphrase, no question mark \\
        \midrule
        Who is the singer who performs \textit{Break it to Me Gently}? & Who is the singer that \textit{gently} performs \textit{Break it to Me}? & Named entity, altered semantics \\
        \midrule
        What is the capital city twinned to Kiel that contains the word "tallinn " in its name & What is the chemical reaction that begins with the letter s & Completely irrelevant paraphrase \\
        \midrule
        What is the official recorded song used during the \textit{British Rule} in Burma? & What is the official recorded melody utilized amid the \textit{British Run the show} in Burma? & Literal paraphrase \\
        \midrule
        who \textit{aircraft} hijacking for immediate cause of War on Terror? & who \textit{flying machine} capturing for quick cause of War on Terror? & Literal paraphrase, poorly formed \\
        \midrule
        Which is the territory that overlaps \textit{Great Britain}? & Which is the domain that covers \textit{Awesome Britain}? & Named entity \\
        \midrule
        What coalition is \textit{Auburn University} a member of? & What fusion is \textit{Reddish-brown College} a part of? & Named entity \\
        \midrule
        The \textit{Cuban Missile Crisis} was a significant event for which countries? & The \textit{Cuban Rocket Emergency} was a critical occasion for which countries? & Named entity \\
        \midrule
        What is U.S. \textit{Gymnastics Hall of Fame} athlete ID of Olga Korbut ? & What is U.S. \textit{Tumbling Corridor of Acclaim} competitor ID of Olga Korbut & Named entity, no question mark \\
        \midrule
        Does the degree of relation of a \textit{great grandfather} equal 3.6? & Does the degree of connection of a \textit{incredible granddad} break even with 3.6? & Literal paraphrase \\
        \midrule
        What is family name of \textit{Neil Diamond} ? & What is family title of \textit{Neil Precious stone} ? & Named entity \\
        \midrule
        What is a afflicts of \textit{red} blood cell? & What could be a torments of \textit{ruddy} blood cell? & Literal paraphrase \\
        \midrule
        What are the based on and the influenced by of Linux? & What is Linux based on and influenced by? & Poorly formed question\\
        \midrule
        Where is the grave that commemorates \textit{Ulysses S. Grant}? & Where is the grave that commemorates \textit{Ulysses S. Allow}? & Named entity \\
        \bottomrule
    \end{tabular}
    \caption{A selection of particularly poor quality data points from \lcq. Data points are copied from the dataset verbatim.}
    \label{tab:appendix-lcq-items}
\end{table*}

Table \ref{tab:appendix-lcq-items} shows a selection of questions and their reference paraphrase taken verbatim from the dataset, along with their main source of error. While there is not always a single source of error, the most serious error for each example is shown.

\end{document}